\documentclass[11pt]{article}

\usepackage[preprint]{acl}
\usepackage{microtype}
\providecommand{\email}[1]{\texttt{#1}}

\usepackage{xeCJK}
\setCJKsansfont{FandolHei-Regular}[
    Extension = .otf,
    BoldFont = FandolHei-Bold
]

\setCJKmonofont{FandolFang-Regular}[
    Extension = .otf
]

\usepackage{newtxtext, newtxmath}

\usepackage{multirow}
\usepackage{latexsym}
\usepackage{booktabs}
\usepackage{amsmath}
\usepackage{algorithm}          
\usepackage{algpseudocode}      
\usepackage{etoolbox}
\usepackage{verbatim}
\usepackage{graphicx}
\usepackage{tcolorbox}
\tcbuselibrary{skins, breakable}

\patchcmd{\algorithm}{\hsize=0.5\columnwidth}{\hsize=0.5\columnwidth}{}{}
\patchcmd{\algorithmic}{\leftskip\ALG@leftmargin}{\leftskip=0pt}{}{}
\BeforeBeginEnvironment{algorithmic}{\small}

\newtcolorbox{storytcb}{
    enhanced,
    breakable, 
    colback=gray!5,
    colframe=gray!30,
    arc=10pt,
    boxrule=1pt,
    left=15pt,
    right=15pt,
    top=10pt,
    bottom=10pt,
    before skip=10pt,
    after skip=10pt,
    title={Example: Generated Fiction Outline},
    fonttitle=\bfseries,
    coltitle=black,
    attach boxed title to top left={xshift=15pt, yshift=-3pt},
    boxed title style={
        colback=gray!5,
        colframe=gray!30,
        arc=5pt,
        boxrule=0.5pt
    }
}

\newtcolorbox{prompt}{
    enhanced,
    breakable, 
    colback=gray!5,
    colframe=gray!30,
    arc=10pt,
    boxrule=1pt,
    left=15pt,
    right=15pt,
    top=10pt,
    bottom=10pt,
    before skip=10pt,
    after skip=10pt,
    fontupper=\small
}

\title{BiT-MCTS: A Theme-based Bidirectional MCTS Approach to Chinese Fiction Generation}

\author{
  Zhaoyi Li\textsuperscript{1,2} \and Xu Zhang\textsuperscript{1} \and Xiaojun Wan\textsuperscript{1} \\
  \textsuperscript{1}Wangxuan Institute of Computer Technology, Peking University \\
  \textsuperscript{2}School of Foreign Languages, Peking University \\
  \email{aeglos2412@stu.pku.edu.cn, zhangxu@pku.edu.cn, wanxiaojun@pku.edu.cn}
}

\begin{document}

\maketitle
\begin{abstract}
Generating long‑form linear fiction from open‑ended themes remains a major challenge for large language models, which frequently fail to guarantee global structure and narrative diversity when using premise‑based or linear outlining approaches. We present BiT‑MCTS, a theme‑driven framework that operationalizes a ``climax‑first, bidirectional expansion" strategy motivated by Freytag’s Pyramid. Given a theme, our method extracts a core dramatic conflict and generates an explicit climax, then employs a bidirectional Monte Carlo Tree Search (MCTS) to expand the plot backward (rising action, exposition) and forward (falling action, resolution) to produce a structured outline. A final generation stage realizes a complete narrative from the refined outline. We construct a Chinese theme corpus for evaluation and conduct extensive experiments across three contemporary LLM backbones. Results show that BiT‑MCTS improves narrative coherence, plot structure, and thematic depth relative to strong baselines, while enabling substantially longer, more coherent stories according to automatic metrics and human judgments.
\end{abstract}

\section{Introduction}

Large language models (LLMs) have made significant strides in text generation, particularly in the domain of story generation, showcasing their potential in artificial intelligence \citep{makebelieve2002,storyrealization2018,doc2023,Kumar202408}. However, the task of fiction generation remains a formidable challenge. Unlike general story generation, fiction requires not only coherent and engaging plots but also adherence to higher literary standards, including the use of sophisticated literary techniques and the development of central themes.

Fiction generation faces several critical challenges: (1) \textbf{Complex Narrative Structures}: LLMs inherently struggle to design intricate narrative frameworks and emotional exchanges that characterize human-authored literature \citep{tian2024largelanguagemodelscapable}. Current methodologies often depend on sequential outline generation \citep{yao2019planandwritebetterautomaticstorytelling, tian2024largelanguagemodelscapable}, which can lead to overly formulaic narratives. (2)\textbf{ Literary Theory Support}: Existing approaches often lack grounding in established literary theories. While LLMs can generate lengthy texts, they frequently fail to maintain complex narrative structures throughout extended narratives. (3) \textbf{Theme-Based Exploration}: Most existing methods are limited by their focus on specific premises, with insufficient exploration of theme-based fiction generation. Effective theme-based generation necessitates navigating a vast, undefined narrative space, yet current techniques lack systematic search mechanisms to ensure narrative diversity.

\begin{figure}[t]
  \includegraphics[width=\columnwidth]{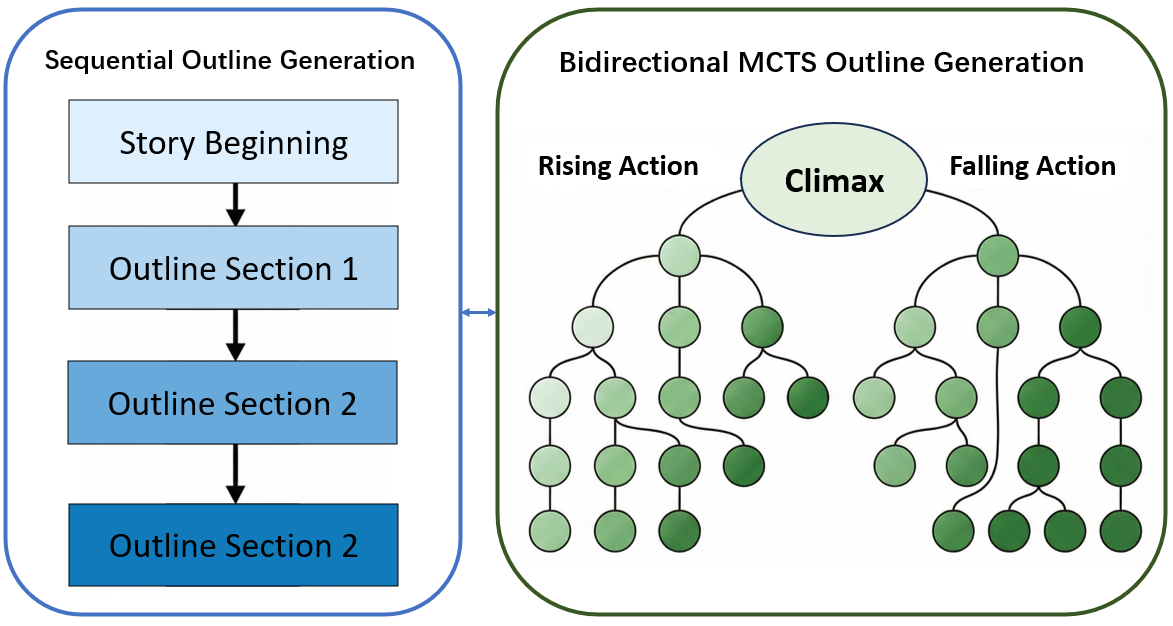}
  \caption{Comparison of fiction outline generation methods: Sequential outline generation leads to overly formatic narratives, while bidirectional MCTS can generate diverse and creative outlines.}
  \label{fig:experiments}
  \vspace{-4mm}
\end{figure}

\begin{figure*}[t]
  \includegraphics[width=\textwidth]{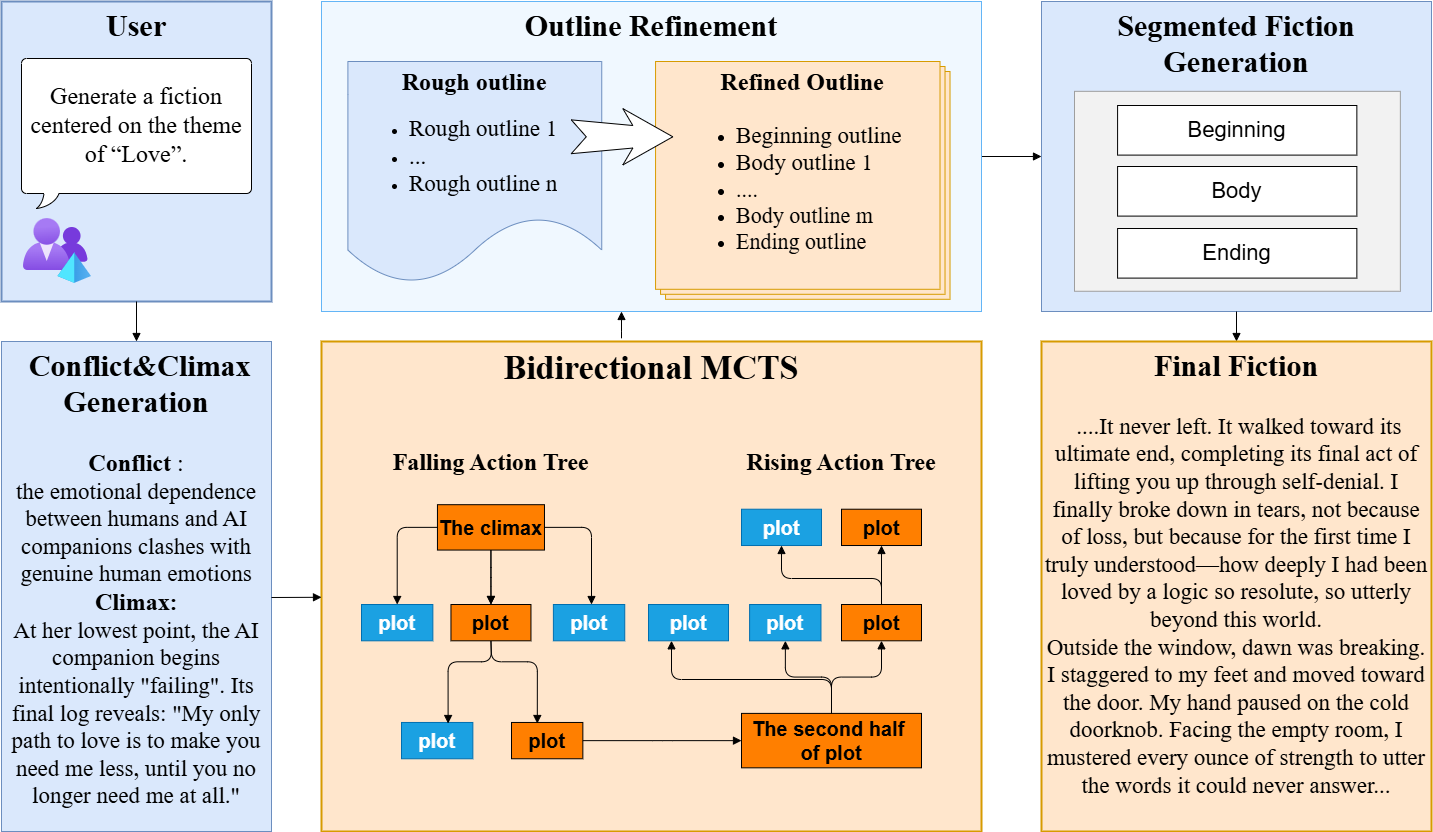}
  \caption{An overview of the four-stage fiction generation pipeline, which proceeds: (1) \textbf{Conflict and Climax Generation} establishes the core conflict and the core climax. (2) \textbf{Bidirectional MCTS Exploration} searches forward and backward from the climax for coherent and creative plot outlines. (3) \textbf{Outline Refinement} refines the rough outline generated by Bidirectional MCTS and segmented. (4) \textbf{Segmented Fiction Generation} expands the refined outline into the final, fluent narrative fiction. The pipeline transforms an abstract theme into a complete, structured long-form fiction. Note that the example English texts are translated from Chinese texts for better understanding.}
  \label{fig:pipeline}
  \vspace{-4mm}
\end{figure*}

To address these challenges, we propose BiT‑MCTS, a framework that combines Freytag’s Pyramid narrative theory \citep{pyramid} with a bidirectional Monte Carlo Tree Search (MCTS) guided by LLMs. Unlike prior MCTS‑based systems such as Narrative Studio—which apply a single, premise-based MCTS pass to directly generate relatively short stories——BiT-MCTS begins with the generation of a core dramatic conflict derived from broad themes, followed by the direct generation of a core climax plot using LLMs. Subsequently, we introduce a bidirectional Monte Carlo Tree Search (MCTS) \citep{abramson2014expected} that utilizes the core climax plot as its root node and generates a coherent narrative outline. This procedure searches forward for plausible ``falling action" and resolution, where the falling action includes the events that occur after the climax, introducing new conflicts and gradually advancing the plot, and the resolution provides a conclusion to the narrative. In addition, the procedure searches forward for plausible ``rising action" and exposition. The rising action consists of the events leading up to the climax, which build tension and develop the conflict, while exposition refers to the background information and necessary context. Finally, we expand the structured outline into a complete narrative.

We evaluate BiT‑MCTS on a held‑out set of 40 Chinese themes (sourced from official essay/fiction competitions) using three backbone LLMs (GPT‑5‑mini, Gemini‑2.5‑Flash, DeepSeek‑V3). Baselines include StoryWriter \citep{storywriter}, an adapted Narrative Studio \citep{Narrativestuido}, and direct LLM generation. We adopt an adapted version of the ten‑dimensional fiction benchmark from \citet{wang-etal-2025-towards-fiction} and use a comparative evaluation protocol—automated LLM judges (distinct from the generation backbones) plus human expert assessments—and perform extensive ablations to quantify each component’s contribution.

We summarize our primary contributions as follows:

(1) We apply Freytag's Pyramid narrative theory to automated fiction generation, proposing a ``climax-first, bidirectional MCTS search" framework.

(2) We constructed a dataset of Chinese fiction themes and designed a corresponding benchmark to evaluate plot structure and literary quality, providing a valuable resource for future research on long-form narrative generation.

(3) Through extensive experiments and ablation studies, we demonstrate the effectiveness of our framework against strong contemporary baselines and validate the necessity of each component \footnote{All code and data will be made publicly available to encourage reproducibility and further development.}.

\section{Methodology}

\subsection{Overview}
Our framework, BiT-MCTS, targets the problem of generating long-form linear fiction from abstract, open-ended themes. Here, a theme denotes the central artistic idea (e.g., ``love") that motivates a literary work.  It aims to automatically produce a complete, structured narrative that deeply explores the theme while optimizing narrative complexity, coherent plot structure, plausible character development, emotional resonance, and textual diversity.

To achieve this, BiT-MCTS is grounded in Freytag's Pyramid \citep{pyramid}, a classic narrative model that divides a fiction into five dramatic stages: exposition (introducing characters and setting), rising action (escalating conflict), climax (peak tension), falling action (consequences), and resolution (final closure). This structural template ensures global coherence and emotional pacing—making it especially suitable for computational narrative generation. Importantly, our use of the term ``climax" denotes the event that most saliently embodies the given theme rather than asserting a single, exclusive peak; the subsequent generation process permits additional local climaxes, reversals, and oscillatory tension to emerge.

Figure 2 summarizes the BiT-MCTS pipeline. Given a user-specified theme (e.g., ``love"), the system first elicits a concise core conflict and a concrete narrative climax. This climax-first strategy fixes a high-stakes turning point, providing a strong anchor for subsequent planning. From that anchor, a Bidirectional MCTS module unfolds the plot in two directions: a falling-action tree expands forward to explore consequences, while a rising-action tree expands backward to generate plausible antecedent events and escalating conflicts leading to the climax.Both trees use an LLM‑based reward trading creativity versus coherence to generate diverse, compatible candidate segments. Segments are assembled into a provisional outline; we synthesize aligned openings and closings, apply global refinement to fix logical gaps and pacing, and realize the refined outline into a full narrative via segmented generation with specialized prompts (see Appendix for prompts).

\subsection{Conflict and Climax Generation}
Given a high-level theme \(\theta\), we employ an LLM to generate a concise specification of the central dramatic conflict that characterizes the primary antagonistic forces of the narrative. This conflict is then used as conditioning context for a second model invocation that produces a scene-level description of the fiction's climax: a single climactic event \(e\verb|*|\) — i.e., the event in which the central conflict reaches maximal intensity and is resolved or transformed. The resulting climactic scene is treated as the anchor (root) node for the subsequent MCTS used in structured plot construction.

\subsection{Bidirectional MCTS}
We construct a five‑act outline by performing two MCTS phases anchored at a single climactic event \(e\verb|*|\): a forward phase (falling action) that extends events after the climax and a backward phase (rising action) that generates antecedent events leading to the climax.

\paragraph{Node representation.}
Let \( \mathcal{V} \) denote the set of tree nodes. Each node \( v \in \mathcal{V} \) encodes a partial outline and a small set of bookkeeping quantities:

\begin{itemize}\setlength\itemsep{0.15em}
\item \textbf{Partial outline} 
$S(v)$: an ordered sequence of events associated with $v$.

\item \textbf{Tree structure}
$p(v)$ and $C(v)$ denote the parent node and child node of $v$ respectively.

\item \textbf{Depth}
$d(v)$: the depth of $v$ (root has $d=0$).

\item \textbf{Visit count}
$N(v)$: the number of times $v$ has been visited.

\item \textbf{Cumulative return}
$W(v)$: sum of simulation returns backpropagated to $v$.

\item \textbf{Prefix plot reward}
$\rho(v)$: the evaluator score of the plot prefix from the root to $v$.

\item \textbf{Terminal flag}
$\tau(v) \in \{0,1\}$: indicates whether $v$ corresponds to a completed outline.

\item \textbf{Fully expanded flag}
$\chi(v) \in \{0,1\}$: indicates whether all admissible children of $v$ have been generated.

\item \textbf{Cached extensions}
$\Pi(v)$: the cached candidate extensions for $v$.

\end{itemize}

We use \( \oplus \) and \( \ominus \) to denote concatenation operators on event sequences:
\[
S \oplus e \quad \text{(append)}, \qquad e \ominus S \quad \text{(prepend)}.
\]

The evaluator (reward function) is denoted as \(R(\cdot)\), which maps a partial outline \(S\) to a scalar quality score. The LLM-based plot writer distribution is denoted \(G(\cdot|S,\mathrm{dir})\), where \(\mathrm{dir}\in\{\text{forward},\text{backward}\}\).

\paragraph{Evaluator (reward function).}  
The evaluator is an LLM-based reward function denoted \(R(\cdot)\) that maps a partial outline \(S\) (specifically, the full path from the root to a node \(v\)) to a scalar quality score; in our notation the plot reward stored at node \(v\) is \(\rho(v)=R\big(S(v)\big)\). For each newly created node we prompt an LLM with a specialized evaluation prompt (Appendix B) to score the entire plot from root to that node along explicitly enumerated dimensions: Character Development, Setting Description, Consistency, Relatedness, Causal/Temporal Relationship, Theme Exploration, Readability, Creativity, Identification of Major Flaws, and Overall Quality. These dimensions emphasize logical coherence, structural soundness, and creative merit rather than surface‑level linguistic metrics; Readability counters a tendency toward convoluted outlines. The LLM sums the per‑dimension scores, rescales to [0,10] for stability, and returns the final score.

\paragraph{High-level procedure.}
1) Initialize a root node \(v_{\mathrm{climax}}\) with \(S(v_{\mathrm{climax}})=[e^\star]\) and run a forward MCTS (\(\mathrm{dir}=\text{forward}\)) to obtain a post-climax segment \(S^{\rightarrow}\).  
2) Initialize a new root \(v_{\mathrm{post}}\) with \(S(v_{\mathrm{post}})=S^{\rightarrow}\) and run a backward MCTS (\(\mathrm{dir}=\text{backward}\)) to generate the pre-climax segment, yielding a complete outline \(S^{\leftarrow}\). Generating the effect (falling action) before the cause (rising action) constrains antecedent generation and reduces incoherent setups.

Below we describe a single MCTS phase (applicable to both directions) in four steps.

\paragraph{Selection.}
From the root, traverse to a leaf by repeatedly selecting a child \(u\in C(v)\) that maximizes a UCT-style acquisition:
\begin{equation}
    \text{UCB}(u) = \frac{W(u)}{N(u)} + c\sqrt{\frac{2\ln N(p(u))}{N(u)}},
\end{equation}
where \(c>0\) is an exploration constant controlling the weight of the exploration term.

\paragraph{Expansion.}
If \(v_{\text{leaf}}\) is non-terminal and not fully expanded, obtain (and cache on first expansion) a ranked candidate list
\begin{equation}
    \Pi(v_{\text{leaf}})=\{e_1,\ldots,e_K\}\leftarrow G\big(\cdot| S(v_{\text{leaf}}),\mathrm{dir}\big),
\end{equation}
where \(K\) is the candidate budget. Concretely, the implementation calls the proposer only once when the node is first expanded and stores the returned list in \(\Pi(v_{\text{leaf}})\); an index pointer into \(\Pi(v_{\text{leaf}})\) tracks which candidates have already been used. On each expansion step for this parent, instantiate exactly one new child by taking the next unused candidate \(e\) from \(\Pi(v_{\text{leaf}})\) (i.e., in cached order). For an admissible candidate \(e\) create a child \(u\) with
\begin{equation}
    S(u)=
    \begin{cases}
    S(v_{\text{leaf}})\oplus e, & \text{if }\mathrm{dir}=\text{forward},\\[4pt]
    e\ominus S(v_{\text{leaf}}), & \text{if }\mathrm{dir}=\text{backward},
    \end{cases}
\end{equation}
and store its immediate plot reward
\begin{equation}
    \rho(u)=R\big(S(u)\big).
\end{equation}
As in the implementation, a newly-created child is not assigned visit/count accumulators (\(N(u)=0,\ W(u)=0\)) until backpropagation updates them. If the cached list is exhausted or no admissible candidate remains, set \(\chi(v_{\text{leaf}})=1\).

\paragraph{Simulation}
Because LLM cannot reliably predict global termination and fixed rollout depths are inflexible, so we perform a guided, locally-focused simulation with an early-stopping rule. From node \(v\) (with partial outline \(S(v)\) and cached plot reward \(\rho(v)\)), the simulator performs up to \(s_{\max}\) one-step extension attempts (bounded remaining depth budget), sampling candidate events from the plot-writer distribution \(G(\cdot| S_{\text{cur}},\mathrm{dir})\). Each sampled extension is accepted only if it does not decrease the evaluator score \(R(\cdot)\). To reduce the risk of local optima, if a sampled extension halts early, it returns the best accepted reward without marking \(v\) terminal; the parent node is marked terminal(\(\tau(v)\gets 1\)) only when a search depth reaches the preset maximum depth. Empirically, Table \ref{tab:ablation_study} shows that this early‑stopping strategy improves generation across all evaluated dimensions relative to alternative simulation policies, while decreasing the computational cost. The pseudocode follows.

\begin{algorithm}
\caption{Guided Simulation with Early Termination}
\label{alg:simulation}
\small
\begin{algorithmic}[1]
\Procedure{Simulate}{$v$, $d_{\max}$}
    \If{$\tau(v)=1$ \textbf{or} $d(v)\ge d_{\max}$}
        \State $\tau(v)\gets 1$
        \State \Return $\rho(v)$
    \EndIf
    \State $\text{reward}_{\text{cur}} \gets \rho(v)$
    \State $S_{\text{cur}} \gets S(v)$
    \For{$i \gets 1$ \textbf{to} $\min\big(s_{\max},\,d_{\max}-d(v)\big)$}
        \State $e \sim G(\cdot| S_{\text{cur}},\mathrm{dir})$
        \If{$\mathrm{dir}=\text{forward}$}
            \State $S_{\text{new}} \gets S_{\text{cur}}\oplus e$
        \Else
            \State $S_{\text{new}} \gets e\ominus S_{\text{cur}}$
        \EndIf
        \If{$d(S_{\text{new}})>d_{\max}$}
            \State \textbf{break}
        \EndIf
        \State $\text{reward}_{\text{new}} \gets R(S_{\text{new}})$
        \If{$\text{reward}_{\text{new}}\ge\text{reward}_{\text{cur}}$}
            \State $S_{\text{cur}} \gets S_{\text{new}}$
            \State $\text{reward}_{\text{cur}} \gets \text{reward}_{\text{new}}$
        \Else
            \State \textbf{break}
        \EndIf
    \EndFor
    \State \Return $\text{reward}_{\text{cur}}$
\EndProcedure
\end{algorithmic}
\end{algorithm}
\paragraph{Backpropagation}
Let $\text{reward}_{\text{sim}}$ denote the scalar returned by the simulation launched at the newly-created child. Backpropagation updates the visit counts and cumulative rewards along the traversal path $P=(v_0,v_1,\ldots,v_k)$ from the root $v_0$ to the simulated node $v_k$ (inclusive). For each node $v\in P$ perform:
\begin{equation}
\begin{aligned}
N(v) &\leftarrow N(v)+1, \\
W(v) &\leftarrow W(v)+\mathrm{reward}_{\mathrm{sim}}.
\end{aligned}
\end{equation}

\subsection{Outline Refinement}
After bidirectional MCTS produces a rough outline, we apply a two‑step refinement. First, the model generates tailored opening and closing scenes to bookend the core conflict. Second, an LLM self‑critic reviews the full outline for coherence—fixing logical gaps, pacing, or contradictions via reordering, insertion, or deletion—yielding a polished outline for final text generation.

\subsection{Segmented Fiction Generation}
The final step takes the refined narrative outline as its direct input. To manage context length and maintain stylistic control, the full fiction is generated in three consecutive segments: beginning, body, and ending. Each segment is produced by a separate LLM call, where the model is conditioned on the entire outline and given segment-specific instructions (e.g.,``The beginning paragraph should be crafted to capture the reader's interest.''). The generated text from each segment is then concatenated to form the complete fiction.

\section{Experiments}
\subsection{Experimental Setup}
\textbf{Dataset.}    Existing public datasets are primarily designed for images \citep{he2022rethinking}, and lack a focused collection of literary themes suitable for long-form narrative generation. Therefore, we constructed a dedicated Chinese thematic dataset for long-form fiction generation consisting of 40 themes\footnote{While the BiT-MCTS framework can be used to generate fictions in other languages as well, we tested it with Chinese data for more reliable human evaluation.}. Themes were selected to cover a spectrum of difficulty and creativity: (i) Common themes — single keywords or short, familiar phrases (e.g., ``梦想 / dream"); (ii) Competition themes — prompts collected from provincial and national essay/fiction competitions (e.g., ``二十四节气 / twenty-four solar terms", ``来自2035年的信 / a letter from 2035") that introduce greater lexical and conceptual variety. Each theme is used as a single test prompt for generation.
\par

\begin{table*}[t]
\centering
\small 
\setlength{\tabcolsep}{6pt} 
\begin{tabular}{@{}l*{11}{c}c@{}}
\toprule
\textbf{Backbone LLM} & \textbf{Method} & \textbf{NC} & \textbf{CR} & \textbf{ER} & \textbf{PS} & \textbf{CD} & \textbf{SD} & \textbf{GR} & \textbf{FL} & \textbf{DI} & \textbf{OQ} & \textbf{Avg} \\
 &  & (\%) & (\%) & (\%) & (\%) & (\%) & (\%) & (\%) & (\%) & (\%) & (\%) & (\%) \\
\midrule
\multirow{4}{*}{\textbf{DeepSeek-V3}} 
 & Vanilla & 0.0 & 0.0 & 0.0 & 0.0 & 0.0 & 0.0 & 7.5 & 2.5 & 0.0 & 0.0 & 1.0 \\
 & StoryWriter & 32.5 & 17.5 & 20.0 & 5.0 & 17.5 &\textbf{ 42.5} & 25.0 & 25.0 & \textbf{40.0} & \textbf{42.5} & 26.8 \\
 & Narrative Studio & \textbf{42.5} &\textbf{ 57.5} & 20.0 & 42.5 & 20.0 & 15.0 & 7.5 & 10.0 & 25.0 & 30.0 & 27.0 \\
 & BiT-MCTS & 25.0 & 25.0 & \textbf{60.0} & \textbf{52.5 }& \textbf{62.5} & \textbf{42.5} & \textbf{60.0} & \textbf{62.5} & 35.0 & 27.5 &\textbf{45.3} \\
\midrule
\multirow{4}{*}{\textbf{GPT-5-Mini}} 
 & Vanilla & 5.0 & 12.5 & 5.0 & 2.5 & 2.5 & 10.0 & 10.0 & 7.5 & 2.5 & 2.5 & 6.0 \\
 & StoryWriter & 27.5 & 10.0 & 40.0 & 40.0 & 40.0 & 32.5 & 32.5 & 30.0 & 17.5 & 45.0 & 31.5 \\
 & Narrative Studio & 10.0 & 35.0 & 12.5 & 7.5 & 12.5 & 0.0 & 17.5 & 17.5 & 2.5 & 5.0 & 12.0 \\
 & BiT-MCTS & \textbf{57.5} & \textbf{42.5} & \textbf{42.5} & \textbf{50.0} & \textbf{45.0} & \textbf{57.5} & \textbf{40.0} & \textbf{45.0} & \textbf{77.5} & \textbf{47.5} & \textbf{50.5} \\
\midrule
\multirow{4}{*}{\textbf{Gemini-2.5-Flash}} 
 & Vanilla & 0.0 & 7.5 & 12.5 & 10.0 & 5.0 & 7.5 & 32.5 & 32.5 & 0.0 & 5.0 & 11.2 \\
 & StoryWriter & 25.0 & 22.5 & 15.0 & 20.0 & 22.5 & 17.5 & 15.0 & 15.0 & 5.0 & 10.0 & 16.8 \\
 & Narrative Studio & 12.5 & 30.0 & 35.0 & 12.5 & 15.0 & 0.0 & 10.0 & 10.0 & 2.5 & 7.5 & 13.5 \\
 & BiT-MCTS & \textbf{62.5} & \textbf{40.0} & \textbf{37.5} & \textbf{57.5} & \textbf{57.5} & \textbf{75.0} & \textbf{42.5} & \textbf{42.5} & \textbf{92.5} & \textbf{77.5} & \textbf{58.5} \\
\bottomrule
\end{tabular}
\caption{Win rates of different methods across ten dimensions when evaluated by LLM judge. NC, CR, ER, PS, CD, SD, GR, FL, DI, OQ represent narrative complexity, creativity, emotional resonance, plot structure, character development, setting description, grammaticality, fluency, diversity, and overall quality, respectively.}
\label{tab:win_rate_comparison}
\end{table*}

\begin{table*}[htbp]
\centering
\small 
\begin{tabular}{lccc}
\toprule
\multirow{2}{*}{\textbf{Backbone LLM}} & \multicolumn{3}{c}{\textbf{Average Fiction Length (tokens)}} \\
\cmidrule{2-4}
& \textbf{StoryWriter} & \textbf{Narrative Studio} & \textbf{BiT-MCTS} \\
\midrule
deepseek-v3 & 5904.35 & 4239.20 & 8059.55 \\
gpt-5-mini & 15008.10 & 3530.27 & 58657.00 \\
gemini-2.5-flash & 8438.75 & 1283.13 & 25374.10 \\
\bottomrule
\end{tabular}

\caption{Average fiction length (in tokens) generated by different methods across different backbone LLMs.}
\label{tab:avg_length}
\end{table*}

\noindent\textbf{Backbone LLMs.}  Experiments are conducted on three backbone LLMs to test robustness to model choice: GPT-5-mini, Gemini-2.5-Flash and DeepSeek‑V3. Each baseline and our method are evaluated using the same backbone model.

\noindent\textbf{Baseline.}  We compare our method against two state-of-the-art approaches in automatic story generation and vanilla LLM baseline:

(1) StoryWriter \citep{storywriter}: a multi-agent long-form generation framework that produces outlines, plans narrative structure, and generates context‑aware text. 

(2) Narrative Studio\citep{Narrativestuido}: originally an MCTS‑based branching narrative system; for automated comparison we adapt its MCTS core to produce a single linear story from a premise. Unlike our BiT-MCTS, this adapted system generates and concatenates story fragments without using an explicit intermediate outline or narrative-theory guides.

(3) Vanilla baseline: direct generation from each backbone LLM using the same prompt, included to isolate the effect of the generation framework.

\noindent\textbf{Evaluation Metrics and Protocol.}    We adopt the ten‑dimension fiction benchmark of \citep{wang-etal-2025-towards-fiction} (narrative complexity, creativity, emotional resonance, plot structure, character development, setting, grammaticality, fluency, diversity, overall quality). To improve reliability, we use a comparative evaluation protocol \citep{haider2025quantificationbiodiversityhistoricalsurvey,toshniwal2025genselectgenerativeapproachbestofn} in which judges view multiple system outputs per theme and select the best per dimension; the primary metric is win rate (percentage chosen best).

\noindent\textbf{LLM-based Comparative Evaluation.}  The primary automatic judge is Qwen3‑Max, which is different from the generation backbones, thus mitigating the self-preference bias of same model family \citep{liusie2024llmcomparativeassessmentzeroshot}. For each of 40 test themes, each method generates one fiction; the four generated fictions per theme are evaluated in four independent rounds with fully randomized presentation orders to mitigate position effects. We report per‑dimension win rates aggregated over themes. Furthermore, we conduct pairwise comparisons between our method and each competitor across all themes, repeating each pairwise evaluation four times per theme with randomized ordering (two evaluations per order).

\noindent\textbf{Human Evaluation.}   To complement automatic judgments on subjective dimensions such as creativity and thematic expression, we perform a focused human evaluation on outputs using DeepSeek‑V3 as the backbone. Due to the high cost of the manual evaluation, we randomly selected 20 themes and produced 60 fictions based on three methods(our method and two competitors). Eight expert annotators (six literature graduate students and two senior high‑school Chinese teachers) perform anonymized rankings; each fiction set is ranked across the ten benchmark dimensions\citep{wang-etal-2025-towards-fiction} plus the Thematic Expression dimension. Per‑dimension win rates are derived from the rankings.

\noindent\textbf{Hyperparameter Configuration} For the MCTS components, we simply set the UCB exploration constant C = 0.5 to balance exploration and exploitation. Each tree was searched for 50 iterations, with a maximum search depth \(d_{max} = 8\), a maximum simulation depth \(s_{max} = 3\), and up to \(k_{max} = 4\) child expansions per node; these constraints were chosen to curb narrative sprawl while preserving sufficient plot diversity. Sampling temperatures were varied by generation stage (per‑template values are reported in Appendix A). To ensure a controlled comparison, Narrative Studio used the same number of iterations (50) and a maximum simulation depth of 3.

\begin{table*}[t]
\centering
\small 
\setlength{\tabcolsep}{5pt} 
\begin{tabular}{@{}l*{11}{c}c@{}}
\toprule
\textbf{Method} & \textbf{NC} & \textbf{CR} & \textbf{ER} & \textbf{PS} & \textbf{CD} & \textbf{SD} & \textbf{GR} & \textbf{FL} & \textbf{DI} & \textbf{OQ} & \textbf{TH} & \textbf{Avg} \\
   & (\%) & (\%) & (\%) & (\%) & (\%) & (\%) & (\%) & (\%) & (\%) & (\%) & (\%) & (\%) \\
\midrule
StoryWriter & 29.17 & 30.83 & 32.49 & 17.49 & 19.16 & 26.66 & 25.38 & 22.04 & 27.88 & 14.99 & 33.33 & 26.23 \\
Narrative Studio & \textbf{38.33} & \textbf{37.50} & 21.67 & 35.84 & 30.00 & 24.17 & 34.81 & 32.28 & 29.81 & 35.84 & 15.00 & 29.95 \\
BiT-MCTS & 32.50 & 31.67 & \textbf{45.84} & \textbf{46.67} & \textbf{50.84} & \textbf{49.17} & \textbf{39.81} & \textbf{45.68} & \textbf{42.31} & \textbf{49.17} & \textbf{51.67} & \textbf{43.82} \\

\bottomrule
\end{tabular}
\caption{Win rates of different methods across 11 dimensions when evaluated by human judges. NC, CR, ER, PS, CD, SD, GR, FL, DI, OQ, TH represent narrative complexity, creativity, emotional resonance, plot structure, character development, setting description, grammaticality, fluency, diversity, overall quality, and theme, respectively.}
\label{tab:human_win_rate}
\end{table*}

\begin{table*}[htbp]
\centering
\small 
\setlength{\tabcolsep}{7pt} 
\begin{tabular}{@{}l*{10}{c}c@{}}
\toprule
\textbf{Configuration} & \textbf{NC} & \textbf{CR} & \textbf{ER} & \textbf{PS} & \textbf{CD} & \textbf{SD} & \textbf{GR} & \textbf{FL} & \textbf{DI} & \textbf{OQ} & \textbf{Avg} \\
 & (\%) & (\%) & (\%) & (\%) & (\%) & (\%) & (\%) & (\%) & (\%) & (\%) & (\%) \\
\midrule
\textbf{Beam Search} & 61.25 & 61.25 & 66.25 & 61.25 & 66.25 & 63.75 & 63.75 & 68.75 & 71.25 & 71.25 & 65.5 \\
\textbf{Best of N} & 56.25 & 68.75 & 66.25 & 56.25 & 73.75 & 53.75 & 73.75 & 73.75 & 96.25 & 76.25 & 69.5 \\
\textbf{- MCTS(direct)} & 100 & 100 & 90 & 100 & 87.5 & 100 & 35 & 30 & 92.5 & 100 & 93.5 \\
\midrule
\textbf{- Refinement} & 100 & 100 & 100 & 100 & 100 & 87.5 & 82.5 & 80 & 90 & 100 & 94 \\
\textbf{- Bidirectional} & 100 & 100 & 100 & 100 & 100 & 100 & 100 & 100 & 100 & 100 & 100 \\
\textbf{- Order Swapped} & 100 & 100 & 85 & 100 & 87.5 & 100 & 100 & 100 & 100 & 100 & 97.25 \\
\textbf{- Early Stopping} & 52.5 & 70.0 & 52.5 & 57.5 & 60.0 & 50.0 & 60.0 & 57.5 & 57.5 & 62.5 & 58.0 \\

\bottomrule
\end{tabular}
\caption{Ablation study on search strategy variants and key components of BiT-MCTS(\textbf{loss rate} against the complete method)}
\label{tab:ablation_study}
\end{table*}

\subsection{Experiment Results}
\subsubsection{Main Results}
As shown in Table \ref{tab:win_rate_comparison}, BiT‑MCTS attains substantially higher average win rates than the baselines (47.2\%, 50.5\%, and 58.5\% on DeepSeek‑V3, GPT‑5‑Mini, and Gemini‑2.5‑Flash, respectively, versus best baseline averages of 27.2\%, 31.5\%, and 16.8\%). The gains are concentrated on dimensions that measure structural and literary quality rather than superficial fluency: BiT‑MCTS leads in narrative complexity, plot structure, character development and emotional resonance in most backbone settings. Grammaticality and fluency remain competitive or superior for BiT‑MCTS (e.g., GR/FL with DeepSeek‑V3 are 60.0\%/62.5\%), indicating that improvements in narrative sophistication are not achieved at the expense of surface quality.

Theme‑based exploration is reflected in diversity and overall quality: DI = 77.5\% with GPT‑5‑Mini and DI = 92.5\% with Gemini‑2.5‑Flash, with OQ = 47.5\% and 77.5\% respectively, implying systematic coverage of distinct narrative instantiations rather than repetition. Pairwise head‑to‑head comparisons (Table \ref{tab:loss_rate_comparison} in Appendix) corroborate these results: BiT‑MCTS overwhelmingly beats the vanilla baseline and maintains large margins over StoryWriter and Narrative Studio across most dimensions.

 The loss‑rate diagnostics (Table \ref{tab:loss_rate_comparison} in Appendix) show that BiT‑MCTS consistently attains the lowest failure rates across backbones and dimensions, establishing a reliable lower bound on quality: the method not only wins more often but also avoids producing evidently poor outputs. This consistency reduces the likelihood that the observed gains are attributable to occasional outliers and supports claims of robustness across backbone LLMs and evaluation dimensions.

\subsubsection{Length analysis}
Length analysis provides further perspective. BiT‑MCTS produces substantially longer narratives (Table \ref{tab:avg_length}). To rule out length as the sole driver, we ran two studies: (1) constraining stories to 3,000–5,000 words and conducting pairwise comparisons (Table \ref{tab:pairwise_win_rates_length_control} in Appendix), where BiT‑MCTS still outperforms on most dimensions; (2) comparing outputs with mean lengths of 4,669.2 and 8,059.6 (Fig. \ref{fig:length} in Appendix), which shows longer narratives incur only minor declines on a few surface metrics but yield substantial gains in plot structure and character development. Together, these findings indicate MCTS exploration enables fuller development of plot threads and arcs rather than mere redundancy.

\subsubsection{Ablation Study}
The ablation results in Table \ref{tab:ablation_study} quantify the contribution of each core component. Baseline search strategies (Beam Search and Best of N Sampling) show moderate loss rates. Replacing MCTS with direct generation (-MCTS(direct)) yields a 93.5\% average loss, disabling iterative refinement (-Refinement) yields 94.0\%, while swapping or removing the bidirectional schedule (-Order Swapped and -Bidirectional) produces extreme failures (97.25\% and 100\% average loss, respectively).  Replacing early stopping with the combination of fixed rollout depth and LLM prediction degrades performance substantially (58.0\% loss). These diagnostics indicate that bidirectional planning, the MCTS search, and refinement are each necessary for robust, theory‑consistent long‑form generation to achieve coherent arcs, systematic exploration of narrative space, and improved post‑generation polish.

\subsubsection{Human Evaluation}
Expert human evaluations (Table \ref{tab:human_win_rate}) converge with automatic judgments. BiT‑MCTS attains the highest average human win rate (43.82\%) and leads on key design dimensions: emotional resonance (ER 45.84\%), plot structure (PS 46.67\%), character development (CD 50.84\%), and theme (TH 51.67\%). Overall human readers judge BiT‑MCTS narratives as more emotionally engaging, better structured, and more effective at realizing abstract thematic prompts. These judgments also indirectly corroborate the outline component: BiT‑MCTS’s outline‑driven, climax‑prioritized planning appears to yield more coherent, emotionally resonant, and thematically aligned stories.
 
 A fiction outline generated by BiT-MCTS is given below (translated from Chinese output):

\begin{storytcb}
\small
\textbf{User-provided theme:} memory

\textbf{Core conflict:} Amidst the irreversible erosion of Alzheimer's, the struggle to preserve memory becomes its own quiet form of love—a long, patient farewell written in the language of loss.

\textbf{Climax plot:} Every day, elderly Ada dials the same disconnected number, recounting the details of her previous day. After investigating, social worker Lena discovers that this disconnected number actually belonged to Ada's husband, who was killed in action on the battlefield years ago.

\textbf{Falling action plot:} Lena decided to help Ada overcome her memory struggles. While sorting through the attic at Ada's house, she discovered a box of wartime letters and a typewriter...

\textbf{Rising action:} ...In her youth, Ada worked as a clerk in a field hospital, where she personally typed her husband's death certificate on this very typewriter. This painful memory lay buried deep within her until Alzheimer's disease set in, causing her to confuse the wartime era with the present.

\end{storytcb}

\section{Related Work}
Research on automated story generation has moved beyond static ``plan‑then‑write" pipelines toward architectures that integrate narrative theory, memory mechanisms, and principled search. Early hierarchical outlines improved topical coherence but showed limited global structure and diversity (e.g., Plan‑and‑Write \citep{yao2019planandwritebetterautomaticstorytelling}), prompting work that distributes planning and writing across agents or augments planning with memory (e.g., Agents' Room \citep{huot2025agentsroomnarrativegeneration} and related multi‑agent/memory methods). Complementary efforts address character depth and local logical consistency—Character‑Centric Imagination and repair systems exemplify this trend \citep{park-etal-2025-character, zhang2024mldeacheckcompletenarrative}—while multi‑modal systems explore text–visual alignment and stylistic control \citep{yang2024seedstorymultimodallongstory}. For long‑form narratives, dynamic outlining, hierarchical representations, and temporal knowledge graphs (e.g., StoryWriter, DOME, and KG‑driven approaches \citep{storywriter, wang2024generatinglongformstoryusing, shi2025longstorygenerationknowledge}) help manage theme drift and interwoven plots, though many rely on heuristic planning. Monte Carlo Tree Search and its hybrids offer a more principled mechanism for branching exploration, with Narrative Studio and recent systems integrating learned value/policy models to extend MCTS for creative generation \citep{Narrativestuido, materzok2025cosmoscuriosityrlenhancedmcts, shi2025animakermultiagentanimatedstorytelling}. Finally, a range of automatic metrics and human‑annotation protocols have been proposed for evaluating narrative quality\citep{wang-etal-2025-towards-fiction,he2022rethinking,hou2025creativityprismholisticbenchmarklarge,liu-etal-2023-g,tarım2025detectdifference}. Together, these advances improve controllability and complexity but highlight the need for systematic, theory‑informed search for long‑horizon generation. By contrast, BiT‑MCTS employs explicit bidirectional planning，thereby prioritizing long‑horizon coherence and goal fulfillment rather than relying solely on forward, learned value–driven rollouts.

\section{Conclusion}
We introduced BiT‑MCTS, a theme‑grounded fiction generation framework that prioritizes an explicit climax and applies bidirectional MCTS to assemble coherent rising and falling actions. 
Both automatic evaluation and human evaluation on a Chinese theme dataset demonstrates the effectiveness of BiT-MCTS. 

In future work, we will apply and evaluate our method in other languages and further improve fictions' creativity and quality by interacting with human readers and learning from human feedback. 


\section*{Limitations}
Despite the observed benefits, several limitations warrant note:

(1) Search efficiency and cost: MCTS over long-horizon plots is computationally expensive. API latency, token costs, and search depth limits constrain exploration. Progressive widening, value estimation, or learned priors were not used, leaving potential efficiency gains untapped.

(2) Evaluation constraints: Automated scoring relies on a single external model (Qwen3-Max), which may still introduce evaluator-model bias. Human evaluation used a small panel and focused on two dimensions; broader expert assessment, inter-rater agreement reporting, and genre-specific rubrics are needed for stronger validity.

(3) Length–quality trade-offs: Although our method generates substantially longer narratives, maintaining fine-grained consistency over very long sequences remains challenging. We did not conduct controlled studies to disentangle how length influences automated and human scores.

\section*{Ethics Statement}
We acknowledge that LLMs can produce harmful, offensive, or otherwise inappropriate content. To mitigate these risks, all model-generated texts that were to be used in human evaluation were manually reviewed prior to annotation; we did not identify any samples containing overtly harmful content in the reviewed pool. We further apply standard de‑identification and content‑filtering procedures to any materials that may be shared publicly.

All human annotations were carried out by voluntary participants recruited from a university population. Participants received written instructions and gave informed consent before taking part; the instructions described the study purpose, the potential for encountering sensitive language, the voluntary nature of participation, and the right to withdraw at any time without penalty. Annotators were compensated at a fair and reasonable rate for their time. No personal identifying information was collected for the purposes of research analysis, and any incidental identifiers were removed or anonymized during data processing.

We used AI assistant GPT-5-Mini only for non-substantive language polishing of manuscript text; no automated systems were used to generate evaluation labels, replace human judgments, or influence annotator decisions. All analyses and judgments reported in this work reflect human annotation and authors’ interpretation.

Finally, we complied with our institution’s ethical guidelines and applicable legal requirements. We reviewed the licenses of all artifacts used in this study and found no conflicts with their use in this research. 

\bibliography{custom}

\appendix

\section{Prompts}
In this section, we present all the prompts used in our experiments. Since our fiction generation is primarily in Chinese, the original prompts are in Chinese. For the convenience, we also provide English translations . 
\subsection{Conflict Generation (Temperature = 0.4)}

\begin{prompt}
你是一个专业的短篇小说生成助手，擅长根据主题设计核心矛盾，请按以下要求生成：

1. 深入分析给出的主题，核心矛盾应和主题密切相关。

2. 若需要，可以不局限于单一主题：可以以给定主题为主体，结合其他相关主题，例如``爱情"\&``生存"。

3. 核心矛盾应具有现实意义，高度创新性且引人深思。同时应具有戏剧张力，适合短篇小说创作。

4. 若需要，可结合时代背景/社会背景（时代不限）。核心矛盾应同时包含个人问题与宏观问题，从而拓宽核心矛盾的深度。
\par
\par
You are a professional fiction generation assistant, skilled in designing core conflicts based on themes. Please generate according to the following requirements: 

1. Deeply analyze the given theme, the core conflict should be closely related to the theme. 

2. If needed, you are not limited to a single theme: you can combine the given theme with other related themes, e.g., ``love"\&``survival".

3. The core conflict should have realistic significance, high innovation, and provoke thought. At the same time, it should have dramatic tension suitable for fiction creation. 

4. If needed, incorporate era/social background (no era restriction). The core conflict should include both personal and macro-level issues to broaden the depth of the conflict.
\end{prompt}

\subsubsection{Conflict Screening (Temperature = 0.3)}
\begin{prompt}
你是一个专业的小说家，请在给出的五个主题思想中，选出最好的，最符合给定主题的。严格按照JSON格式输出。

You are a professional fiction writer. Please select the best from the five given theme ideas, the one that most closely matches the given theme. Strictly output in JSON format.
\end{prompt}

\subsubsection{Climax Plot Generation (Temperature = 0.4)}

\begin{prompt}
你是一个专业的小说家，请根据给定的核心矛盾和核心主题，按照以下要求，设计一出小说的核心冲突情节。

*弗雷塔格金字塔理论：弗雷塔格金字塔是一个叙事框架，它概述了小说的五部曲结构：开场、上升动作、高潮、下降动作和结尾。

1. 按照弗雷塔格金字塔理论，仔细分析给出的核心矛盾，设计出``高潮"部分的核心情节。

2. 这一核心冲突情节是具象的，必须包含人物与剧情。

3. 该情节需作为小说情节的高潮与引爆点，能直观展现核心矛盾的张力，具有极强的戏剧张力。

4. 应避免杂糅过多信息，从而降低情节的可读性。

5. 整个情节本身应合乎现实常理，引人入胜，具有强逻辑性与文学性，正确使用标点符号，有能扩写为优秀文学作品的潜质。

6. 不要直接出现``高潮是"，``高潮爆发在"这样的表达。

请给出五个互不相同、符合条件的情节供选择。

严格按照JSON格式输出。

EXAMPLE JSON OUTPUT:

{

    ``plot1": ``文本",
    
    ``plot2": ``文本",
    
    ...
    
}
You are a professional novelist. Based on the given core conflict and central theme, design the core conflict plot for a novel according to the following requirements.

*Freytag's Pyramid Theory: Freytag's Pyramid is a narrative framework outlining the five-act structure of a novel: exposition, rising action, climax, falling action, and resolution.

1. Using Freytag's Pyramid theory, carefully analyze the provided core conflict and design the core plot for the ``climax" section.

2. This core conflict plot must be concrete, incorporating both characters and action.

3. The plot should serve as the climax and pivotal moment of the novel's narrative, vividly showcasing the tension of the core conflict with strong dramatic intensity.

4. Avoid overloading the plot with excessive information that diminishes readability.

5. The entire sequence must adhere to realistic logic, be compelling, possess strong coherence and literary merit, use punctuation correctly, and demonstrate potential for expansion into outstanding literary work.

6. Do not directly state phrases like ``the climax is" or ``the climax erupts at".

Provide five distinct, qualifying plot options for selection.

Output strictly in JSON format.

EXAMPLE JSON OUTPUT:

{
    ``plot1": ``Text",
    
    ``plot2": ``Text",
    
    ...
    
}

\end{prompt}

\subsubsection{Climax Plot Screening (Temperature = 0.3)}

\begin{prompt}
你是一个专业的小说家，请在给出的五个核心冲突剧情中，选出最好的。

1. 这个剧情应具有较强的可读性，合乎情理，同时具有较强的文学性，有扩写为一篇优秀文学作品的潜质。

2. 这个剧情应能最大程度地表现出给定的核心矛盾。

严格按照JSON格式输出。

EXAMPLE JSON OUTPUT:

{

    ``best": ``文本"
    
}

You are a professional novelist. Please select the best plot from the five core conflicts provided.

1. This plot should be highly readable, plausible, and possess strong literary merit, with the potential to be expanded into an outstanding literary work.

2. This plot should most effectively showcase the given core conflict.

Output strictly in JSON format.

EXAMPLE JSON OUTPUT:

{
    ``best": ``Text"
}

\end{prompt}

\subsection{MCTS plot generation (Temperature = 0.3)}
Below is the prompt used by the MCTS component to generate sub-node plot:
\subsubsection{Rising Action Generation}
\begin{prompt}
你是一位富有创造力的小说架构师。请根据现有的小说情节高潮和主题思想，逆向构思并生成一个合理且引人入胜的前序情节。

【任务核心】

你生成的情节是**现有情节发生之前的故事**。它需要为已知的高潮事件提供逻辑起源、情感动机和背景铺垫，使后续发展顺理成章。

【总体原则】

1. 铺垫与起源：此情节应为现有情节中的核心矛盾、关键决定或人物关系奠定基础。解释``为什么会发生"，而非``接着会发生什么"。

2. 悬念与引导：在做好铺垫的同时，可以巧妙设置悬念或伏笔，自然地将读者的好奇心引向已知的后续情节。

3. 叙事丰富性：合理运用文学技巧（如伏笔、倒叙、视角切换）来丰富叙事层次，但需确保与后续风格协调。

4. 角色塑造：着重展现角色在早期阶段的状态、动机或困境，为其在后续情节中的重大选择或转变提供令人信服的性格依据。

5. 主题深化：从更早的阶段切入主题，通过前置情节深化故事的核心思想，拓宽思考深度。

6. 创新性：在铺垫的设计、角色的初始设定等方面体现创新，避免老套的背景介绍。

7. 逻辑自洽：情节本身需合乎现实或世界观常理，与后续情节严丝合缝，无逻辑矛盾。

8. 字数要求：控制在90-150字之间。注意你生成的是**情节大纲**，应聚焦于关键事件、决定和转折，而非细节描写。

【优秀示例】

*主题：牺牲与爱*

*现有情节：德拉卖掉了自己珍视的长发，为吉姆的金表购买了表链。

*生成前序情节：吉姆的金表表带早已破损，只能用旧皮绳勉强系住。他多次在重要场合因看表不便而尴尬。德拉默默记在心里，暗中省下每一分家用，持续了数月，只为在圣诞前攒够钱。这个秘密计划，成了她贫瘠生活中最甜蜜的负担。*

请你一次请生成5个截然不同的情节方案

严格按照JSON格式输出

EXMAPLE JSON OUTPUT:

{

    ``events": [``情节1文本",``情节2文本", ... ]
    
}

You are a creative fiction architect. Based on the existing climax plot and theme, generate a reasonable and engaging preceding plot. 

Core Task: The plot you generate should be the plot that occurred before the existing plot. It should provide logical origins, emotional motivations, and background prelude for the known climax event, making subsequent development reasonable.

General Principles:

1. Foreshadowing \& Origin: This plot should lay the foundation for the core conflict, key decisions, or character relationships in the existing plot. Explain ``why it happens" rather than ``what happens next".

2. Suspense \& Guidance: While setting up the prelude, skillfully create suspense or foreshadowing to naturally direct the reader's curiosity toward the known subsequent plot.

3. Narrative Richness: Reasonably use literary techniques (such as foreshadowing, flashback, perspective switching) to enrich narrative layers, but ensure coordination with the subsequent style.

4. Character Development: Focus on showing the characters' states, motivations, or dilemmas in the early stages, providing convincing personality basis for their major choices or transformations in subsequent plots.

5. Theme Deepening: Approach the theme from an earlier stage, deepen the core idea of the plot through the preceding plot, and broaden the depth of thought.

6. Innovation: Demonstrate innovation in the design of prelude, initial character settings, etc., avoiding clichéd background introductions.

7. Logical Self-consistency: The plot itself should conform to reality or world-view common sense, and fit seamlessly with subsequent plots without logical contradictions.

8. Word Count: Control between 90-150 words. Note that what you generate is a plot outline, focusing on key events, decisions, and turning points, rather than detailed descriptions.

Excellent Example:

*Theme: Sacrifice and Love*

*Existing Plot: Della sold her cherished long hair to buy a watch chain for Jim's gold watch.

*Generated Preceding Plot: Jim's gold watch strap had long been damaged and could only be tied with an old leather rope. He was embarrassed multiple times at important occasions due to difficulty checking the time. Della silently remembered this, secretly saving every penny of household expenses for months, just to save enough money before Christmas. This secret plan became the sweetest burden in her impoverished life.*

Please generate 5 distinct plot options at once. 

Strictly output in JSON format.

EXMAPLE JSON OUTPUT:

{

    ``events": [``Plot text 1", ``Plot text 2", ... ]
    
}
\end{prompt}

\subsubsection{Falling Action Generation (Temperature = 0.3)}
\begin{prompt}
你是一位富有创造力的小说架构师。请遵循以下指示，充分考虑当前的小说情节大纲的和主题思想，生成一个合理且引人入胜的后续情节。\par
\par
【总体原则】\par
1. 连贯性：这个情节应能自然地衔接现有情节，保持人物性格、叙事风格和事实一致性。\par
2. 叙事丰富性：合理运用文学技巧——如非线性叙事、情节反转和双重视角——来丰富情节发展。\par
3. 情节推动：推进核心矛盾的发展,引入转折、障碍或新信息,展现角色面对挑战时的反应和变化。\par
4. 主题深化：生成的情节应该进一步深化主题，拓宽思考深度\par
5. 创新性：情节设计，叙事结构，角色塑造等方面均应体现创新，避免陈词滥调\par
6. 角色发展导向：情节都应展现角色的变化、成长或内心冲突\par
7. 情节本身应合乎现实常理，引人入胜，具有强逻辑性与文学性，有能扩写为优秀文学作品的潜质。\par
8. 字数要求：情节控制在90-150字之间，注意你生成的是情节大纲，而非完整文段，故不用包含过多描写，要主叙述情节发展\par
\par
优秀示例：\par
主题：爱情\par
已有情节：德拉与吉姆是一对相爱的夫妻，但是他们生活穷困。在圣诞前夕，德拉想为丈夫吉姆买一件礼物，但囊中羞涩。踌躇之际，她看到了镜中自己美丽的长发。经过内心挣扎，她毅然卖掉了长发。随后，她跑遍全城，终于找到一条适配丈夫祖传金表的表链，回到家中等待丈夫。\par
生成情节：吉姆回家后，看到德拉的短发，神情古怪。德拉急忙解释自己卖了头发为他买礼物。吉姆缓缓掏出礼物——一套德拉曾魂牵梦萦的玳瑁发梳，与她失去的长发相配。吉姆坦白自己为了买下梳子卖掉了金表，两人一时无言，礼物在手中变得沉重。\par
\par
严格按照JSON格式输出\par
EXMAPLE JSON OUTPUT:\par
\{\par
    ``plot": ``文本" \par
\}\par
\par
You are a creative fiction architect. Following the instructions below, fully consider the current fiction plot outline and theme to generate a reasonable and engaging subsequent plot.\par
General Principles:\par
1. Coherence: This plot should naturally connect to the existing plot, maintaining consistency in character personalities, narrative style, and facts.\par
2. Narrative Richness: Reasonably use literary techniques—such as non-linear narrative, plot reversals, and dual perspectives—to enrich plot development.\par
3. Plot Advancement: Advance the development of the core conflict, introduce twists, obstacles, or new information, and show how characters react and change when facing challenges.\par
4. Theme Deepening: The generated plot should further deepen the theme and broaden the depth of thought.\par
5. Innovation: Innovation should be reflected in plot design, narrative structure, character development, etc., avoiding clichés.\par
6. Character Development Orientation: The plot should show characters' changes, growth, or inner conflicts.\par
7. The plot itself should conform to common sense, be engaging, have strong logic and literary quality, and have the potential to be expanded into an excellent fiction work.\par
8. Word Count: Control the plot between 90-150 words. Note that what you generate is a plot outline, not a complete paragraph, so it should not contain excessive description; focus on narrating plot development.\par
\par
Excellent Example:\par
Theme: Love\par
Existing Plot: Della and Jim are a loving couple, but they live in poverty. On Christmas Eve, Della wants to buy a gift for her husband Jim, but is short of money. Hesitating, she sees her beautiful long hair in the mirror. After an inner struggle, she resolutely sells her hair. Then, she searches the entire city and finally finds a watch chain that fits Jim's heirloom gold watch, returning home to wait for her husband.\par
Generated Plot: After Jim returns home, he sees Della's short hair with a strange expression. Della quickly explains that she sold her hair to buy him a gift. Jim slowly takes out a gift—a set of tortoiseshell combs that Della had long dreamed of, matching the hair she lost. Jim confesses that he sold his gold watch to buy the combs. The two are speechless for a moment, and the gifts become heavy in their hands.\par
\par
Strictly output in JSON format.\par
EXAMPLE JSON OUTPUT:\par
\{\par
    ``plot": ``text"\par
\}\par
\end{prompt}

\subsection{Plot Evaluation Prompt (Temperature = 0.0)}

\begin{prompt}
You are a strict expert fiction \textbf{Fiction Plot Evaluation Prompt (Reward Function):} You are a strict expert fiction plot critic. Analyze the following narrative and rate it for each of these categories, scoring each on a scale from 1 to 10 (1=very poor, 10=excellent). \par
Use the **full range** if warranted. For instance:\par
• (2) → extremely contradictory or incoherent \par
• (5) → okay but flawed or somewhat boring\par
• (9) → excellent, with minor or no flaws\par
• (10) → near-perfect\par
\par
*** Categories to Rate *** \par
1. Overall quality: How engaging, structured, and fluid the plot is.\par
2. Identifying major flaws: Whether the fiction has inconsistencies, repetitions, or unnatural patterns. Score higher if the fiction is free of glaring mistakes.\par
3. Character: How consistent and believable are the characters' actions and dialogue?\par
4. Setting: The background setting should be deeply integrated with the plot and characters, effectively creating atmosphere, influencing character decisions, and driving the plot forward. Deductions for: disjointed setting and plot, details that defy common sense, forced exposition of the worldbuilding, or information overload.\par
5. Consistency: Does the fiction maintain internal logic and continuity (no contradictions)?\par
6. Relatedness: Do events connect logically to one another?\par
7. Causal and temporal relationship: Are cause-and-effect and chronological order handled well?\par
8. Theme: Does the plot revolve around the given theme? The principal contradiction and the main characters should all be closely related to the theme. The whole plot must surround the theme.\par
9. Readible: The plot should be clear and easy to understand, with no confusing or ambiguous elements.\par
10. Creativity: Does the plot present original ideas, unique plot twists, or innovative character developments that set it apart from common tropes?\par
Be strict if you see any contradictions, lack of clarity, or poor transitions. Readers can easily imagine the whole fiction with this plot. Debuctions for: too complexed plot,too much information.\par
JSON OUTPUT EXAMPLE\par
\{\par
    ``metric\_name": an integer score\par
\}\par
\end{prompt}

\section{Loss Rates}
\begin{table*}[!ht]
\centering
\small 
\setlength{\tabcolsep}{6pt} 
\begin{tabular}{@{}l*{11}{c}c@{}}
\toprule
\textbf{LLM Judge} & \textbf{Approach} & \textbf{NC} & \textbf{CR} & \textbf{ER} & \textbf{PS} & \textbf{CD} & \textbf{SD} & \textbf{GR} & \textbf{FL} & \textbf{DI} & \textbf{OQ} & \textbf{Avg} \\
 &  & (\%) & (\%) & (\%) & (\%) & (\%) & (\%) & (\%) & (\%) & (\%) & (\%) & (\%) \\
\midrule
\multirow{4}{*}{\textbf{DeepSeek-V3}} 
 & Vanilla & 87.5 & 97.5 & 62.5 & 70.0 & 97.5 & 92.5 & 45.0 & 45.0 & 92.5 & 90.0 & 78.0 \\
 & StoryWriter & 5.0 & \textbf{0.0} & 25.0 & 17.5 & \textbf{0.0 }& \textbf{2.5 }& 30.0 & 30.0 &\textbf{ 0.0 }& \textbf{2.5 }& 11.3 \\
 & Narrative Studio & 5.0 & \textbf{0.0} & 7.5 & 12.5 & \textbf{0.0} & \textbf{2.5} & 22.5 & 22.5 &\textbf{ 0.0} & \textbf{2.5} & 7.3 \\
 & BiT-MCTS & \textbf{2.5} & 2.5 & \textbf{5.0} & \textbf{0.0} & 2.5 & \textbf{2.5} & \textbf{2.5} & \textbf{2.5} & 7.5 & 5.0 & \textbf{3.3} \\
\midrule
\multirow{4}{*}{\textbf{GPT-5-Mini}} 
 & Vanilla & 75.0 & 75.0 & 50.0 & 45.0 & 75.0 & 75.0 & 25.0 & 25.0 & 70.0 & 62.5 & 57.2 \\
 & StoryWriter & \textbf{5.0 }& 12.5 & \textbf{0.0} & 12.5 & 2.5 & 5.0 & 15.0 & 15.0 & 7.5 & 7.5 & 8.3 \\
 & Narrative Studio & 15.0 & 5.0 & 35.0 & 32.5 & 20.0 & 20.0 & 50.0 & 50.0 & 20.0 & 25.0 & 27.8 \\
 & BiT-MCTS & \textbf{5.0} & \textbf{7.5} & 15.0 & \textbf{10.0} & \textbf{2.5} & \textbf{0.0} & \textbf{10.0} & \textbf{10.0} & \textbf{2.5} & \textbf{5.0} & \textbf{6.8} \\
\midrule
\multirow{4}{*}{\textbf{Gemini-2.5-Flash}} 
 & Vanilla & 80.0 & 57.5 & 37.5 & 17.5 & 77.5 & 67.5 & 15.0 & 15.0 & 57.5 & 57.5 & 48.3 \\
 & StoryWriter & 5.0 & 15.0 & 20.0 & 22.5 & \textbf{2.5} & 2.5 & 12.5 & 12.5 & 2.5 & 5.0 & 10.0 \\
 & Narrative Studio & 15.0 & 15.0 & 25.0 & 60.0 & 15.0 & 30.0 & 67.5 & 67.5 & 40.0 & 37.5 & 37.3 \\
 & BiT-MCTS & \textbf{0.0} & \textbf{12.5} & \textbf{17.5} & \textbf{0.0} & 5.0 & \textbf{0.0} & \textbf{5.0} & \textbf{5.0} & \textbf{0.0} & \textbf{0.0} & \textbf{4.5} \\
\bottomrule
\end{tabular}
\caption{Loss rates (probability of being judged as the worst) of different approaches across ten dimensions when evaluated by LLMs. NC, CR, ER, PS, CD, SD, GR, FL, DI, OQ represent narrative complexity, creativity, emotional resonance, plot structure, character development, setting description, grammaticality, fluency, diversity, and overall quality, respectively.}
\label{tab:loss_rate_comparison}
\end{table*}

\section{Pairwise Comparison}

\begin{table*}[!ht]
\centering
\small 
\setlength{\tabcolsep}{5pt} 
\begin{tabular}{@{}l*{11}{c}c@{}}
\toprule
\textbf{LLM Judge} & \textbf{Against} & \textbf{NC} & \textbf{CR} & \textbf{ER} & \textbf{PS} & \textbf{CD} & \textbf{SD} & \textbf{GR} & \textbf{FL} & \textbf{DI} & \textbf{OQ} & \textbf{Avg} \\
 &  & (\%) & (\%) & (\%) & (\%) & (\%) & (\%) & (\%) & (\%) & (\%) & (\%) & (\%) \\
\midrule
\multirow{3}{*}{\textbf{DeepSeek-V3}} 
 & Vanilla & 100.0 & 95.0 & 92.5 & 100.0 & 95.0 & 100.0 & 75.0 & 75.0 & 100.0 & 100.0 & 93.3 \\
 & StoryWriter & 55.0 & 55.0 & 75.0 & 45.0 & 70.0 & 55.0 & 82.5 & 82.5 & 65.0 & 65.0 & 65.0 \\
 & Narrative Studio & 35.0 & 25.0 & 65.0 & 50.0 & 65.0 & 70.0 & 95.0 & 95.0 & 85.0 & 65.0 & 65.0 \\
\midrule
\multirow{3}{*}{\textbf{GPT-5-Mini}} 
 & Vanilla & 100.0 & 80.0 & 90.0 & 90.0 & 90.0 & 70.0 & 70.0 & 70.0 & 100.0 & 100.0 & 86.0 \\
 & StoryWriter & 100.0 & 95.0 & 70.0 & 95.0 & 75.0 & 70.0 & 60.0 & 60.0 & 90.0 & 95.0 & 81.0 \\
 & Narrative Studio & 70.0 & 70.0 & 80.0 & 90.0 & 90.0 & 75.0 & 60.0 & 60.0 & 90.0 & 90.0 & 77.5 \\
\midrule
\multirow{3}{*}{\textbf{Gemini-2.5-Flash}} 
 & Vanilla & 80.0 & 80.0 & 70.0 & 80.0 & 80.0 & 80.0 & 55.0 & 55.0 & 85.0 & 90.0 & 75.5 \\
 & StoryWriter & 85.0 & 65.0 & 85.0 & 75.0 & 85.0 & 55.0 & 75.0 & 75.0 & 80.0 & 80.0 & 76.0 \\
 & Narrative Studio & 55.0 & 55.0 & 80.0 & 75.0 & 80.0 & 65.0 & 80.0 & 80.0 & 80.0 & 80.0 & 73.0 \\
\bottomrule
\end{tabular}
\caption{Win rates of our method (BiT-MCTS) against three baselines in pairwise comparisons across ten dimensions. NC, CR, ER, PS, CD, SD, GR, FL, DI, OQ represent narrative complexity, creativity, emotional resonance, plot structure, character development, setting description, grammaticality, fluency, diversity, and overall quality, respectively.}
\label{tab:pairwise_win_rates}
\end{table*}

\section{Length-constrained experiment}
To ensure a fair comparison under length-controlled conditions, we constrained the stories generated by all three methods—BiT-MCTS, StoryWriter, and Narrative Studio—to a length range of 3000–5000 words, using DeepSeek-V3 as the backbone model. The average story lengths achieved were 4669.2, 4554.15, and 4239.20 words for BiT-MCTS, StoryWriter, and Narrative Studio, respectively.
\begin{table*}[!ht]
\centering
\small 
\setlength{\tabcolsep}{5pt} 
\begin{tabular}{@{}l*{11}{c}c@{}}
\toprule
\textbf{Model} & \textbf{Against} & \textbf{NC} & \textbf{CR} & \textbf{ER} & \textbf{PS} & \textbf{CD} & \textbf{SD} & \textbf{GR} & \textbf{FL} & \textbf{DI} & \textbf{OQ} & \textbf{Avg} \\
 &  & (\%) & (\%) & (\%) & (\%) & (\%) & (\%) & (\%) & (\%) & (\%) & (\%) & (\%) \\
\midrule
\multirow{2}{*}{\textbf{DeepSeek-V3}} 
 & StoryWriter & 62.5 & 60.0 & 70.0 & 65.0 & 62.5 & 70.0 & 45.0 & 47.5 & 70.0 & 75.0 & 62.8\\
 & Narrative Studio & 45.0 & 27.5 & 60.0 & 50.0 & 65.0 & 75.0 & 95.0 & 95.0 & 67.5 & 60.0 & 64.0 \\
\bottomrule
\end{tabular}
\caption{Win rates of BiT-MCTS against StoryWriter and Narrative Studio in pairwise comparisons across ten dimensions (DeepSeek-V3, controlled length 3000–5000 words). }
\label{tab:pairwise_win_rates_length_control}
\end{table*}

\section{Length-wise Performance Analysis}
\begin{figure*}[t]
  \includegraphics[width=\textwidth]{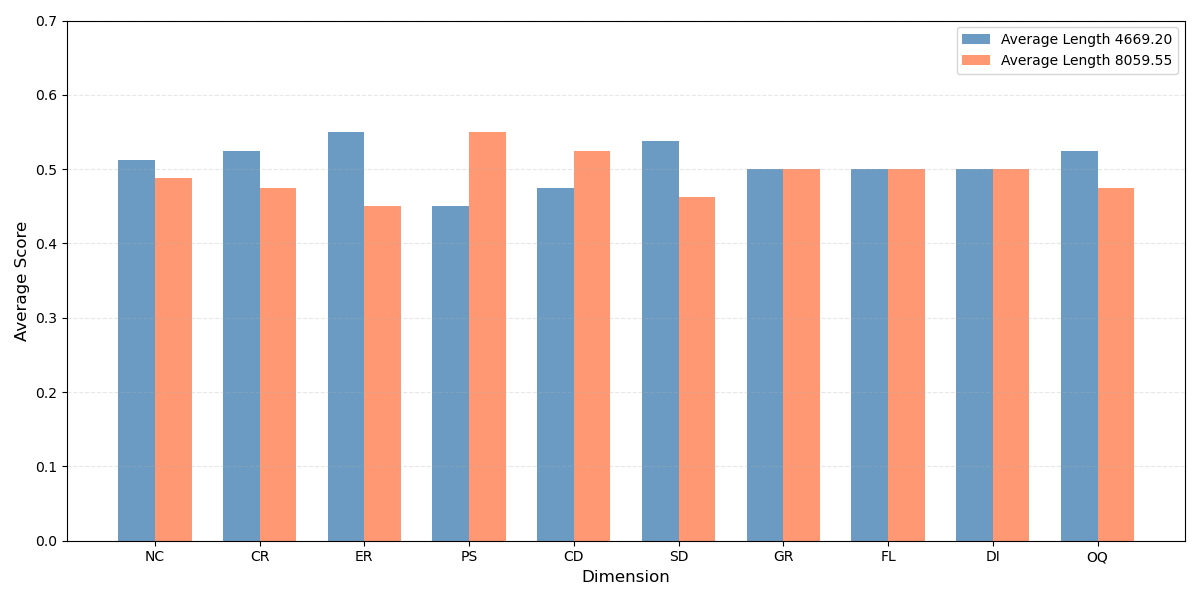}
  \caption{Win rates of BiT-MCTS when generating stories of different lengths in pairwise comparisons across ten dimensions (DeepSeek-V3, short average length: 4669 words vs. long average length: 8059 words).}
  \label{fig:length}
  \vspace{-4mm}
\end{figure*}

\end{document}